\documentclass[10pt, a4paper]{article}
\usepackage{lrec2022} 
\usepackage{multibib}
\newcites{languageresource}{Language Resources}
\usepackage{graphicx}
\usepackage{tabularx}
\usepackage{soul}
\usepackage{titlesec}
\titleformat{\section}{\normalfont\large\bfseries\center}{\thesection.}{1em}{}
\titleformat{\subsection}{\normalfont\SmallTitleFont\bfseries\raggedright}{\thesubsection.}{1em}{}
\titleformat{\subsubsection}{\normalfont\normalsize\bfseries\raggedright}{\thesubsubsection.}{1em}{}
\renewcommand\thesection{\arabic{section}}
\renewcommand\thesubsection{\thesection.\arabic{subsection}}
\renewcommand\thesubsubsection{\thesubsection.\arabic{subsubsection}}

\usepackage{epstopdf}
\usepackage[utf8]{inputenc}

\usepackage{hyperref}
\usepackage{xstring}

\usepackage{color}
\usepackage{arabtex}
\usepackage{utf8}
\usepackage{slashbox}
\setcode{utf8}
\usepackage{makecell}

\usepackage{algorithm}
\usepackage{algpseudocode}
\usepackage{tablefootnote}
\usepackage{xcolor}
\usepackage{amsmath} 
\usepackage{pdfpages}
\usepackage{svg}

\usepackage{array}

\newcolumntype{P}[1]{>{\centering\arraybackslash}p{#1}} 

\usepackage{todonotes}
\usepackage{comment}
\usepackage[normalem]{ulem}

\usepackage{xcolor}
\usepackage{xspace}
\colorlet{kw}{blue}
\definecolor{com}{rgb}{0,0.6,0.3}

\newcommand\quran{Qur’an\xspace}

\usepackage[symbol]{footmisc}
\usepackage[acronym]{glossaries}

\newdimen\origiwspc%
\newdimen\origiwstr%
\origiwspc=\fontdimen2\font
\origiwstr=\fontdimen3\font

\title{
\iftrue
TCE$^\ast$
\else
MatMulMan
\fi
at \quran QA 2022:\\ Arabic Language Question Answering Over Holy \quran Using a Post-Processed Ensemble of BERT-based Models
}

\name{Mohammed ElKomy, Amany M. Sarhan}

\address{Computer and Control Engineering Department, Faculty of Engineering, Tanta University. \\
\{mohammed.a.elkomy, amany\_sarhan\}@f-eng.tanta.edu.eg\\
}

\abstract{
In recent years, we witnessed great progress in different tasks of natural language understanding using machine learning. Question answering is one of these tasks which is used by    search engines and social media platforms for improved user experience.
Arabic is the language of the Holy \quran; the sacred text for 1.8 billion people across the world.
Arabic is a challenging language for Natural Language Processing (NLP) due to its complex structures.
In this article, we describe our attempts at OSACT5 Qur'an QA 2022 Shared Task,
which is a question answering challenge on the Holy \quran in Arabic.
We propose an ensemble learning model based on Arabic variants of BERT models.  In addition, we perform post-processing to enhance the model predictions.
Our system achieves a Partial Reciprocal Rank (pRR) score
of 56.6\% on the official test set.
\fontdimen2\font=0.1ex
\\
\newline
}

\begin{document}
    \maketitleabstract

    \section{Introduction}
    Nowadays, the web and social media are integral parts of our modern digital life as they are the main sources of the unprecedented amounts of data we have.
    Thanks to the breakthrough in deep learning,
    search engines are no longer restricted to keyword matching; instead, they are currently able to understand queries in natural language and satisfy the intended information need of users.
    Question Answering (QA) is an essential task in information retrieval which is forming the basis for the new frontier of search engines.
    Plenty of studies on question answering systems has been performed on English and other languages. However, very few attempts have addressed the problem of Arabic question answering ~\cite{survey}.

    Arabic NLP research in question answering is particularly challenging due to the scarcity of  resources and the lack of processing tools available for Arabic.
    Arabic language, as well, has some unique characteristics by being a highly inflectional and derivational language with complex morphological structures~\cite{albayan}.
    The Holy \quran is the sacred text for Muslims around the globe and it is the main source for teachings and legislation in Islam~\cite{ayatec},
    there are 114 chapters in \quran corresponding to 6,236 verses, every verse consists of a sequence of words in Classical Arabic (CA) dating back to 1400 years ago.

    This paper describes our proposed solutions for OSACT5 Qur'an QA 2022 shared task. The shared task introduced QRCD (The Qur’anic Reading Comprehension Dataset), which is a dataset for extractive question answering.
    First, we experimented with a variety of Arabic pre-trained Bidirectional Encoder Representations from Transformers
    (BERT) models, then we implemented an Ensemble approach to get more robust results from a mixture of experts (MOEs).
    After that, we propose some post-processing operations to enhance the quality of answers according to the official evaluation measures.
    The task is evaluated as a ranking task according to the Partial Reciprocal Rank (pRR) metric.

    The rest of this paper is organized as follows. In section \ref{sec:related}, we describe the related work,
    in section \ref{sec:dataset}, we outline the dataset details and official evaluation measures,
    in section \ref{sec:qa_design}, we explain the system design and the implementation details,
    in section \ref{sec:eval}, we report the system evaluation results,
    and finally section \ref{sec:conc} concludes the paper.
    \footnote{
    The source code and trained models are available at \href{https://github.com/mohammed-elkomy/quran-qa}{\color{blue}https://github.com/mohammed-elkomy/quran-qa}.
    }
    \color{white}
    \footnote{\label{foot}To enable fair comparison among the teams, the organizers only considered 238 examples for the official test-split results and excluded 36 samples due to being very similar to public splits.}\iftrue
    \color{white}
    \renewcommand{\thefootnote}{\fnsymbol{footnote}}
    \footnote[1]{Tanta Computer Engineering}
    \renewcommand{\thefootnote}{\arabic{footnote}}
    \color{black}
    \fi

    \color{black}

    \section{Related Work}
    \label{sec:related}
    Question answering systems have been an active point of research in recent years, particularly for highly-resourced languages such as English.
    ~\cite{squad} introduced SQuAD1.0 dataset which is a widely used dataset for question answering in English.
    To tackle the data scarcity in Arabic NLP, ~\cite{SOQAL} presented Arabic Reading Comprehension Dataset (ARCD) which consists of 1,395 questions posed by crowdworkers.
    Moreover, ~\cite{SOQAL} automatically translated SQuAD1.0 using google translation services.
    Only a little attention has been paid to question answering on \quran.
    ~\cite{albayan} proposed a Support Vector Machine (SVM) question answering system with handcrafted features
    to extract answers from both \quran and its interpretation books (Tafseer).
    Recent work by  ~\cite{ayatec} introduced \texttt{AyaTEC} as the first fully reusable test collection for Arabic QA on the Holy \quran where questions are posed in Modern Standard Arabic
    (MSA) and their corresponding answers are qur'anic verses in CA.

    \section{Dataset and Task Description}
    \label{sec:dataset}
    In this section, we describe QRCD (The Qur’anic Reading Comprehension Dataset),
    the problem definition and official evaluation metrics of the \quran QA 2022 shared task of answering questions on the holy \quran ~\cite{QRCD_repo}.

    \subsection{Dataset Details  }
    {The} Qur'anic Reading Comprehension Dataset
    {(QRCD)~\cite{qrcd} is the first large scale question answering dataset on the holy \quran text. It was introduced as a part of the \quran QA 2022 Shared Task for question answering~\cite{qrcd}.}
    The dataset consists of 1,093 tuples of question-passage pairs { that are coupled with their extractive answers to constitute 1,337 question-passage-answer triplets. The question-passage pairs are} split into training, development and
    testing sets as shown in Table \ref{tab:splits_table}. The dataset follows the same format as the commonly used reading comprehension dataset SQuAD1.0~\cite{squad}.
    However, QRCD is quite different in terms of size as it is much smaller than SQuAD1.0 which contains 100k unique pairs/questions.
    {In addition, unlike SQuAD, the QCRD contains a small number of unique questions, each of which is repeated multiple times with different passage and answer pairs. As shown in Table \ref{tab:splits_table}, } the number of unique questions in QRCD is much lower than the number of question-passage pairs.
    {This poses an additional challenge for learning a question answering system which should be able to predict different answers to the same question under different passage contexts}. 

    \begin{table}[!h]
        \begin{center}
            \begin{tabularx}{\columnwidth}{|l|X|X|X|}
                \hline
                \backslashbox[33mm]{Aspect}{Split} & \textbf{Train} & \textbf{Dev} & \textbf{Test}           \\
                \hline
                Question-passage pairs             & 710            & 109          & 274$^\text{\ref{foot}}$ \\
                \hline
                Unique questions                   & 118            & 17           & 34                      \\
                \hline
            \end{tabularx}
            \caption{Number of question-passage pairs and number of unique questions in each split of QRCD dataset.}
            \label{tab:splits_table}
        \end{center}
    \end{table}

\begin{figure*}[htbp]
        \centering
        \includegraphics[page=1,width=.85\textwidth]{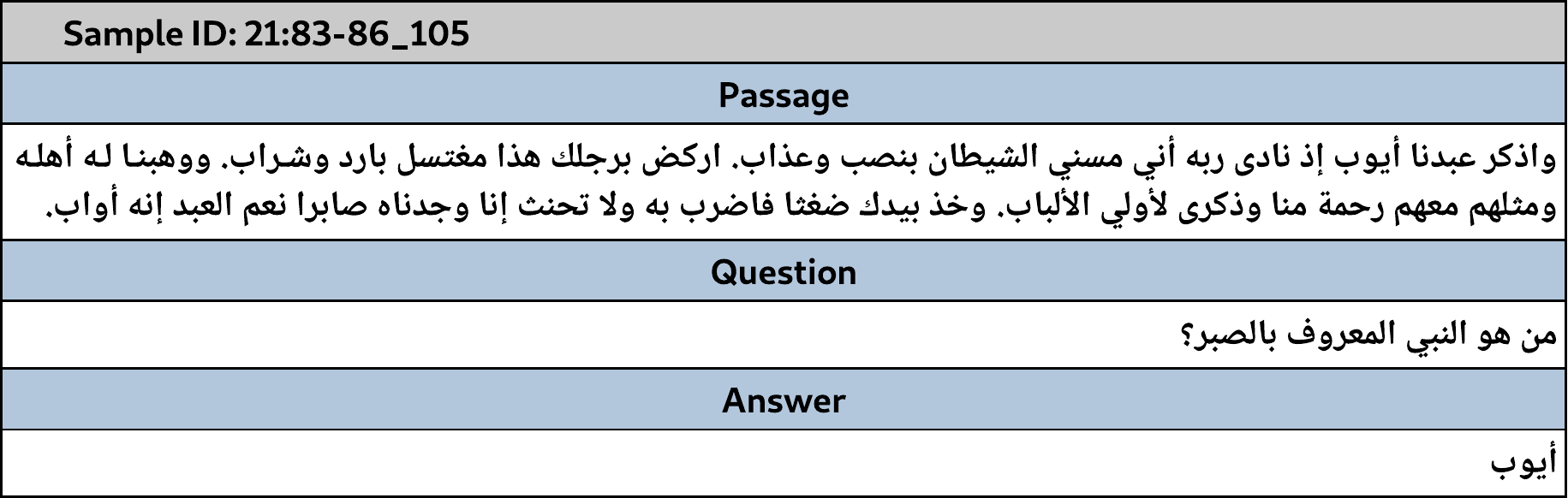}
        \caption{An example of question-passage-answer triplet from QRCD \protect\cite{qrcd}.}
        \label{fig:qrcd_sample}
    \end{figure*}

    The QRCD dataset draws its inspiration from the prior work \texttt{AyaTEC} by reformulating the test collection into an  extractive question answering task ~\cite{qrcd}.
    Each sample in QRCD is a question-passage-answer triplet which comprises a question in MSA,
    a passage taken from the Holy Qur'an
    \footnote{
    The Holy Qur'an is a very special classical Arabic text revealed 1,400 years ago, making it extremely challenging for computational linguistics tasks.
    }
    that spans one or more consecutive verses,
    and an answer to the question extracted from the passage. Figure \ref{fig:qrcd_sample} demonstrates an example from the QRCD dataset\footnote{The verses from the Holy Qur'an in the dataset come from the simple-clean text style (diacritics removed) from
    \href{http://tanzil.net/}{Tanzil Project}.}.

    \subsection{\quran QA Shared Task Description}
    {The \quran QA 2022 shared task~\cite{qrcd} aims to develop models for extractive question answering on the holy \quran passages. Given a Qura'nic passage and a question, the solution to the shared task should extract the answer to the question from the input passage. The answer always exists as a span within the given passage. Questions could be either factoid or non-factoid. Solutions to the shared task are required to extract any correct answer to the input question from the context passage even when the passage has more than one answer.}


    \subsection{{Task Evaluation Measures}}
    \label{subsec:prr_metric}
    {
    This question answering task is evaluated as a \textit{ranking} task. The QA system will return up to 5 potential answers ranked from the best to the worst according to their probability of correctness. The task evaluation measure produces a higher score when the correct answer is ranked at a higher position.
    When the correct answer is predicted among the 5 potential answers but it is at a lower rank, then the evaluation score is discounted. The task adopts the \textit{partial Reciprocal Rank (\textbf{pRR})}~\cite{ayatec} as the official evaluation metric. Partial Reciprocal Rank (\textbf{pRR}) is a variant of the \textit{Reciprocal Rank} (RR) evaluation metric which is a commonly used metric for ranking tasks. Unlike \textit{Reciprocal Rank} (RR), the  \textit{partial Reciprocal Rank (\textbf{pRR})} will give credit to systems that predict answers with a partial inexact matching with the correct answer.
    Equation \ref{eq:prr} formally describes the \textbf{pRR} metric evaluation for a ranked list of answers $A$, where $m_{r_{k}}$ is the partial matching score for the returned answer at the $k^{th}$ rank,
    as $k$ is taken to be equal to the rank position of the first answer with a non-zero matching score. More details can be found at ~\cite{ayatec} 5.2.
    \begin{equation}
        p R R(A)=\frac{m_{r_{k}}}{k} ; k=\min \left\{k \mid m_{r_{k}}>0\right\}
        \label{eq:prr}
    \end{equation}
    In addition to the pRR scores, the task evaluation system reports other metrics such as the Exact Match (\textbf{EM}) and \textbf{F1@1}. The \textbf{EM} score is a binary measure that will be equal to one when the top predicted answer exactly matches the ground truth answer. The \textbf{F1@1} metric measures the degree of token overlap between the top predicted answer and any of the ground truth answers. Scores computed from individual passage-question-answer triplets are averaged to compute the overall score over the entire evaluation dataset.}

    \section{{QA System Design}}
    \label{sec:qa_design}
    {\textbf{B}idirectional \textbf{E}ncoder \textbf{R}epresentations from
    \textbf{T}ransformers (BERT)~\cite{bert} models achieve state-of-the-art results in many natural language understanding problems. Our solution is an ensemble of BERT based models pre-trained on Arabic language corpora and fine-tuned on the shared task dataset. We built an ensemble that merges predictions from individual models. Additionally, we also designed and implemented a set of post-processing operations that aim to improve the quality of predicted answers and boost the task evaluation measure.}
    In this section, we describe our QA system developed to solve the \quran QA 2022 challenge. First, we give a brief background on BERT models and their usage for question answering tasks. Then we provide an overview of Arabic language BERT models that we used to build our ensemble. Finally, we provide the details of our ensemble building approach and the proposed post-processing operations to improve predicted answers quality.

    \subsection{BERT for Question Answering}

    \subsubsection{BERT}
    \label{subsec:bert}
    BERT models achieve their state-of-the-art performance due to a procedure called pre-training which allows BERT to discover the language structures and patterns.
    BERT uses two pre-training tasks, namely masked language model (MLM) and next sentence prediction (NSP)~\cite{bert}.
    After that, a second stage called fine-tuning, which is performed to adapt the model for a downstream task by making use of the features learnt during the pre-training phase.

    \subsubsection{Question Answering using BERT}
    \label{subsec:bert_qa}
    As mentioned in \ref{subsec:bert}, the fine-tuning phase takes a pre-trained model and stacks a randomly-initialized output layer suitable for a particular downstream task.
    For extractive question answering, both the question and passage are tokenized and packed into a single sequence
    and the output layer is required to give a probability for the i$^{th}$ token in the passage to be the start of the answer span $P_i$ using the dot product
    of a start vector $S$ and the i$^{th}$ token's hidden representation $T_i$ as seen from equation \ref{eqn:start_softmax}, a similar analysis holds for the end of the answer span with an end vector $E$.
    The score of a candidate's answer span from the i$^{th}$ token to the j$^{th}$ token is defined in Equation \ref{eqn:span_score}.
    Answers are only accepted for $j \geq i$ since the two probability distributions are independent and not guaranteed to produce a valid span ~\cite{bert}.
    $S$ and $E$ are trainable weights and randomly initialized layers stacked on top of pre-trained BERT.
    In our case the system is not limited to just one answer as in SQuAD1.0 ~\cite{squad}, instead, a ranked list of 20 answers is generated from the model and ranked based on the span score as in Equation \ref{eqn:span_score}.
    \begin{equation}
        P_{i}=\frac{e^{S \cdot T_{i}}}{\sum_{j} e^{S \cdot T_{j}}}
        \label{eqn:start_softmax}
    \end{equation}
    \begin{equation}
        Span_{i,j} = S T_{i} + E T_{j}
        \label{eqn:span_score}
    \end{equation}


    \subsection{Arabic variants of BERT model}
    \label{subsec:models}
    The standard BERT model variants are not pre-trained on Arabic text which hinders the development of Arabic NLP. Nevertheless, various researchers working on the Arabic natural language understanding have developed variants of BERT models that were trained on Arabic corpora. 
    We made use of those models, which are discussed later in this section, to build our ensemble.

    \subsubsection{AraBERT}
    The work by~\cite{arabert} introduced AraBERT model which inherits the exact same architecture from BERT. However, being pre-trained on a large Arabic corpus of 24GB of text
    collected from news articles and Wikipedia dumps.
    In this work, we use \textbf{bert-large-arabertv02} and \textbf{bert-base-arabertv02} available on the huggingface community~\cite{hug}.

    \subsubsection{QARiB}
    QARiB~\cite{qarib} is another Arabic BERT variant pretrained on a mixture of formal and informal Arabic with state-of-the-art support for Arabic dialects and social media text,
    ~\cite{qarib} released 5 BERT models to the community pre-trained on corpora of different sizes. 

    \subsubsection{ARBERT and MARBERT}
    ARBERT and MARBERT~\cite{arbert} are two Arabic-specific Transformer-based MLM pre-trained on a widely large Modern Standard Arabic (MSA) corpus of 61GB of text for the case of ARBERT,
    while MARBERT is pre-trained on 128GB of text focused on both dialectal Arabic (DA) and MSA.

    \subsection{Training Details}
    \label{subsec:training}
    We fine-tune a set of five Arabic BERT models as mentioned in \ref{subsec:models}, namely AraBERT-v02$_\text{Large}$\footnote{A subscript Large and Base refers to the model size.},
    AraBERT-v02$_\text{Base}$, QARiB$_\text{Base}$, ARBERT and MARBERT
    \footnote{ARBERT and MARBERT uses BERT$_\text{Base}$ architecture.}.

    The training objective is to maximize the log-likelihoods of the correct start and end token positions~\cite{bert}.
    For the development phase, we train BERT$_\text{Base}$ and BERT$_\text{Large}$ for 50 and 65 epochs respectively, looking for the epoch at which the model performs best on the validation split.
    For the test phase, we train BERT$_\text{Base}$ and BERT$_\text{Large}$ for 32 and 40 respectively.
    We used a batch size of 8 for BERT$_\text{Large}$ and 16 for BERT$_\text{Base}$ and a learning rate of 2e-5 for all of our models.

    \subsection{Span Voting Ensemble}
    \label{subsec:ensemble}
    In this work, we build an ensemble from several different Arabic BERT models that we discussed earlier in \ref{subsec:training}.
    The ensemble approach is effective for cancelling the noise exhibited by individual models through majority voting among experts. i.e. Mixture of Experts (MOE).
    We treat the answer spans as discrete entities. 
    For each sample, we {consider} the top 20 predictions made by each model along {with its correctness probability. For} each candidate prediction,
    we {compute the sum of its associated correctness probabilities from all models}. We formulate the voting process as follows.
    \vspace{-5pt}$$\alpha_{s,e}=\sum_{j=0}^{M}{\alpha^{j}_{s,e}}$$
    Where $\alpha_{s,e}$ represents the summed ensemble correctness probability for the answer span starting at token $s$ and ending at token $e$,
    and $\alpha^{j}_{s,e}$ represents the correctness probability for the same answer span for the $j^{th}$ expert of the $M$ experts taken into account.
    \\
    {
    After that, the set of all possible answers considered by the ensemble is sorted according to their summed ensemble correctness probabilities.
    Finally, the entire ranked list is post-processed and truncated for only the top 5 answers to be evaluated by \textbf{pRR@5}.
    }


    \vspace{-3pt} \subsection{Post-processing}
    By carefully reviewing the answer spans predicted by the BERT models we fine-tuned, we found some systematic errors causing sub-optimal predictions.
    Here we propose some basic post-processing rules to improve the model predictions.
    The post-processing pipeline takes a ranked list of answer spans, it typically takes at least 20 answer spans from a single model or the span-voting ensemble.
    Figure \ref{fig:post_example} provides an illustrative example of the post-processing pipeline. Due to the limited space, we only consider the top 15 answer spans from the original system outputs.
    \subsubsection{Handling Sub-words}
    \label{subsec:subword}
\begin{figure*}[!t]
        \centering
        \includegraphics[page=1,width=1\textwidth]{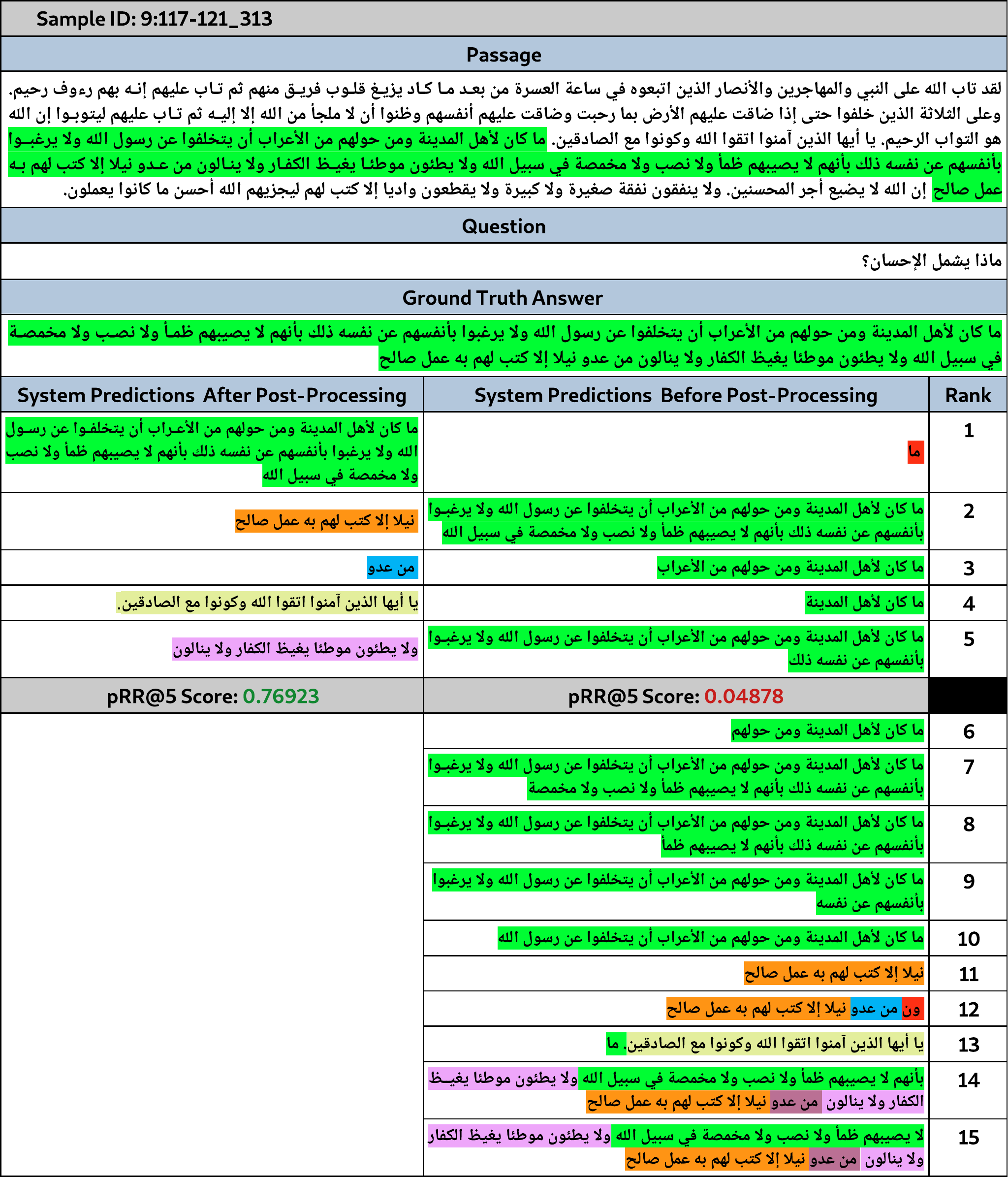}
        \caption{A comprehensive example for post-processing.
        Here we present the outputs of the system before post-processing (original outputs) on the right and after post-processing on the left.
        We used green to highlight the ground truth answer or parts of it extracted by the system. Words and sub-words marked by red are dropped according to the rules \ref{subsec:subword} and
        \ref{subsec:uninfor}.
        Other colours used for highlighting text depict the high overlap among the predictions before post-processing.
        }
        \label{fig:post_example}
    \end{figure*}

    Before feeding the input to BERT, it must undergo the tokenization step.
    Tokenization is the process of splitting a sentence into tokens.
    For BERT, WordPiece tokenizer is commonly used as a subword tokenizer.
    This makes the system susceptible to producing an output with incomplete words like "\RL{وفي الرقاب والغارم}" which will be penalized by the evaluation process,
    \fontdimen2\font=0.80ex
    we perform a simple post-processing rule to extend or drop tokens such that we do not have broken words and
    this simple rule produces a corresponding output \\
    "\RL{وفي الرقاب والغارمين}",
 \fontdimen2\font=.6    ex
    for the previously mentioned example. In Figure \ref{fig:post_example}, we dropped the sub-token "\RL{ون}" at rank 12 which is part of "\RL{ينالون}".
     \fontdimen2\font=\origiwspc
    \subsubsection{Redundancy Elimination}
    \label{subsec:no_overlap_post_process}
\begin{figure}[ht]
        \centering
        \includegraphics[page=1,width=.5\textwidth]{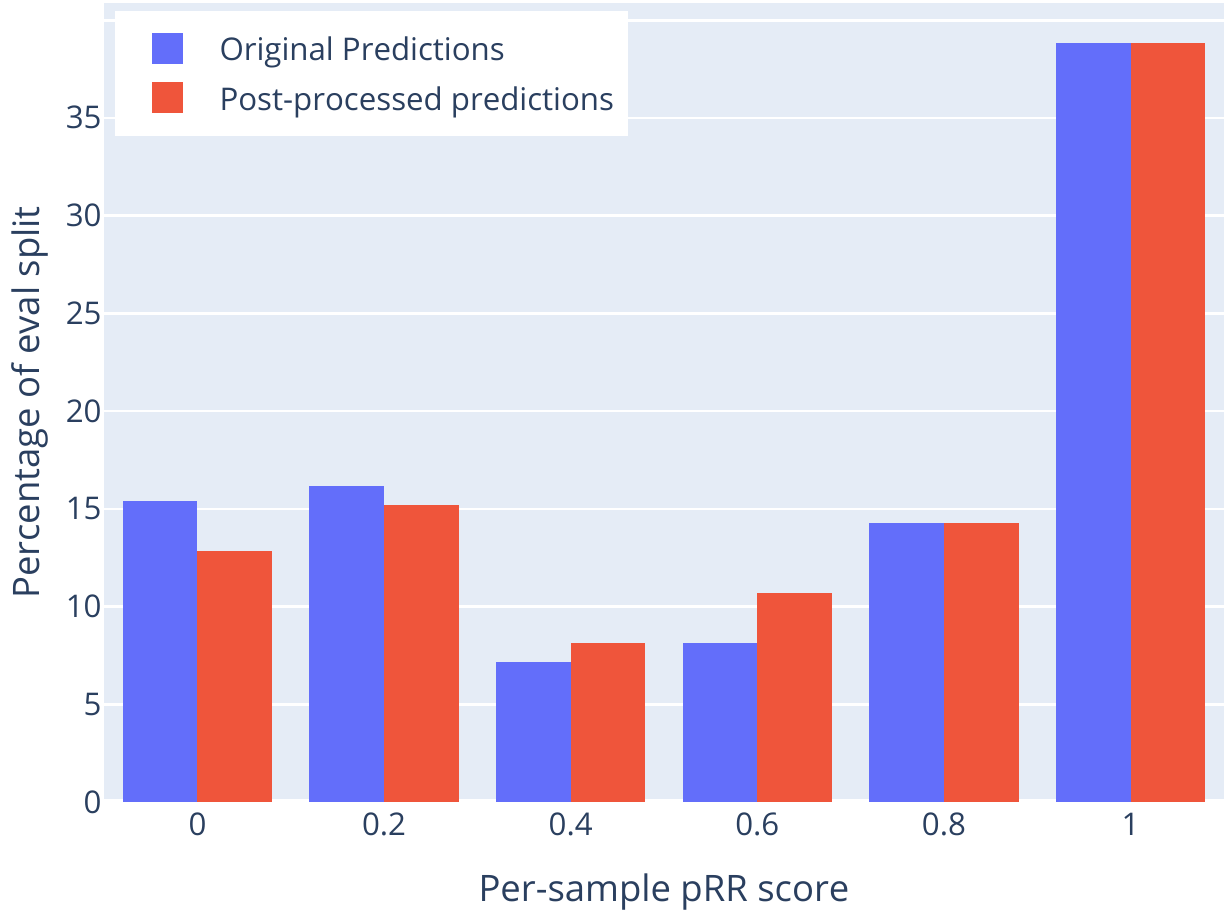}
        \caption{The post-processing impact on the per-sample pRR score distribution, for this plot, we used the ouputs of the models marked with $\dagger$ in Table \ref{table:eval}.}
        \label{fig:histo}
    \end{figure}

    We analyzed the predictions of a variety of our fine-tuned models and discovered that most of the answer spans are highly overlapping with each other,
    making the ranked list suboptimal with respect to the pRR metric in \ref{subsec:prr_metric}.
    The rationale behind this is, the pRR metric \textit{only} considers the (k+1)$^{th}$ prediction
    when there is no overlap with any of the ground-truth answer tokens for the k$^{th}$ prediction,
    This implies repeating any of the words from the k$^{th}$ prediction in the (k+1)$^{th}$ prediction is suboptimal, which is a common behaviour exhibited by BERT for QA.
    Here we present algorithm \ref{alg:no_overlap} which ensures the elimination of span overlap among the answers of a ranked list returned by the system.
    An illustrative example showing the predictions before and after the application of this rule is given in Figure \ref{fig:post_example},
    the colours used for highlighting text depict the high overlap between unprocessed answers, it clearly shows the span "\RL{ما كان لأهل المدينة}" is common in the first few unprocessed answers.    \fontdimen2\font=.70001ex
    After applying this rule, the ranked list after post-processing better covers the text in the passage and the pRR score increases from $0.048$ to $0.769$  after post-processing.
    Figure \ref{fig:histo} shows the distribution of the per-sample pRR score before and after post-processing,
    the percentage of development samples of the first two bins after post-processing is reduced, which means we are less observing completely wrong answers after post-processing.
    \fontdimen2\font=\origiwspc
    \begin{algorithm}[!t]

        \caption{Redundancy Elimination Algorithm}\label{alg:no_overlap}
        \hspace*{10pt}\textbf{Input}: $P$, passage text; $A$, input answers list. \\
        \hspace*{10pt}\textbf{Output}: post-processed answers list.
        \begin{algorithmic}
            \State $N_{P} \gets$  Number of words in $P$.
            \State $M_{seen} \gets$ $0_{N_P}$ \Comment{Initialized to Zero}\\
            \Comment{The mask used to track seen words}
            \State $A_{post} \gets$: [] \Comment{Output Initialized to an empty list}\\

            \For{\textbf{each} $a$ \textbf{in} $A$}
                \State $s$  $\gets$ get start-word index of span $a$.
                \State $e$    $\gets$ get end-word index of span $a$.
                \State $a_{seen}$ $\gets$ $M_{seen}$[$s$:$e$]  \\\Comment{Get seen slice of answer span $a$}
                 \If
                {$a_{seen}$ has any zero}\\
                  {
                     \Comment{at least a word is not seen,}\\
                     \Comment{marked by 1 in $a_{seen}$}
                     \State $a_{\text{ unseen of } P}$ $\gets$ get\_unseen\_seqs($a_{seen}$,$P$)\\
                     \Comment{brings unseen contiguous sequences of words.}

                      \For{\textbf{each} $seq_\text{unseen}$ \textbf{in} $a_{\text{ unseen of } P}$}\\
                             \Comment{$seq_\text{unseen}$ is a subsequence of words   }\\
                             \Comment{with $a_{seen}$ consisting of only zeros  }
\fontdimen2\font=0.8ex
                            \State $s_{unseen}$ $\gets$ start-word index of $seq_\text{unseen}$.
                            \State $e_{unseen}$ $\gets$ end-word index of $seq_\text{unseen}$.
                            \State $seq_{text}$ $\gets$ text spanned by $seq_\text{unseen}$.

    \fontdimen2\font=1ex
                            \\  \hspace*{42pt}  $M_{seen}$[$s_{unseen}$: $e_{unseen}$] = 1
                                \hspace*{10.5pt} \\\Comment{Mark tokens as seen }\\

                           \hspace*{42pt} $A_{post}$ = $A_{post} \smile seq_{text}$
                            \\ \Comment{Append this unseen part of answer span $a$}
                     \EndFor \\
                    }
                \EndIf
            \EndFor
        \end{algorithmic}
    \end{algorithm}

    \subsubsection{Uninformative Answer Removal}
    \label{subsec:uninfor}In this post-processing rule, we remove the uninformative answers from the ranked list, We define an uninformative answer as having one of the following conditions:
    \begin{enumerate}
        \item{
            All of the stemmed answer tokens exist in the stemmed question tokens, for example,  a question like "\RL{ما هي شجرة الزقوم?}" with a complete answer span predicted by the system "\RL{الزقوم}" is considered an uninformative answer.
        }
           \item{The whole answer span consists of stop-words, which can never meet the information need of a question in QRCD, for example, answer spans like "\RL{اذا}", "\RL{ليس}", "\RL{ثم}" are considered uninformative answers.}

    \end{enumerate}

\vspace{-5pt}
    Uninformative answers come from two sources, first, the original output of the BERT model without post-processing
    and second, the post-processed outputs after removing redundant tokens as in \ref{subsec:no_overlap_post_process}.
    For the example in Figure \ref{fig:post_example}, the answer at rank 1 "\RL{ما}" is rejected due to being uninformative.

    \subsubsection{Updating the Ranked List}
    After performing the post-processing pipeline described in \ref{subsec:no_overlap_post_process}, \ref{subsec:uninfor} and \ref{subsec:subword} in order,
    we may end up with a new ranked list with more than 5 answer spans, we only consider the top 5 answer spans in the post-processed ranked list for the metric evaluation (pRR@5).
%

    \section{Experimental Evaluation}
    \label{sec:eval}
     \vspace{-0pt}
    In this section, we report our trained models' results along with the official scores and run details on the Codalab competition.
    \begin{table*}[ht]
        \begin{center}
            \begin{tabular}{|l|P{1.25cm}|P{1.2cm}|P{1.4cm}|}
                \hline
                \textbf{Single Models}              & \textbf{EM} (\%) & \textbf{F1} (\%) & \textbf{pRR} (\%) \\
                \hline
                $^\dagger$arabertv02$_\text{Large}$ & 37.2             & 59.0             & \textbf{61.7}     \\
                $^\dagger$arabertv02$_\text{Base} $ & 36.5             & 58.5             & 60.7              \\
                $^\dagger$ARBERT                    & 37.3             & 58.7             & 60.9              \\\Xhline{5\arrayrulewidth}
                MARBERT                             & 32.1             & 51.5             & 53.9              \\
                QARiB$_\text{Base}$                 & 25.9             & 45.3             & 48.1              \\
                \hline  \multicolumn{4}{l}{  \vspace{-5pt} } \\
                \hline
                \textbf{Ensemble}                   & \textbf{EM} (\%) & \textbf{F1} (\%) & \textbf{pRR} (\%) \\
                \hline
                {Ensemble$_\textbf{Vanilla}$}                          & 39.4             & 59.4             & 63.97 \\
                {{Ensemble$_\textbf{POST}$}}                     & 38.5             & 59.4             & \textbf{65.22}    \\
                \hline
            \end{tabular}
            \caption{
            QRCD development split results, reported metrics for single models are averaged for a number of model checkpoints trained with different seeds,
            while for the ensemble case, it is a single instance produced from combining different checkpoints.
            \textbf{Ensemble$_\textbf{Vanilla}$} refers to combining 45 checkpoints of models indicated by $\dagger$ (15 for each).
            \textbf{Ensemble$_\textbf{POST}$} represents the \textbf{Ensemble$_\textbf{Vanilla}$} output after post-processing.
            }
            \label{table:eval}
        \end{center}
    \end{table*}

    \begin{table*}[ht]
        \begin{center}
            \begin{tabular}{|l|P{1.3cm}|P{1.3cm}|P{1.4cm}|}
                \hline
                \textbf{Run ID}               & \textbf{EM} (\%) & \textbf{F1} (\%) & \textbf{pRR} (\%) \\
                \hline
                Ensemble$_\text{keep}$        & 26.8             & 48.5             & 55.7              \\ 
                Ensemble$_\text{remove}$      & 26.8             & 50.0             & \textbf{56.6}     \\ 
                \hline
            \end{tabular}
            \caption{   Test phase official results on QRCD dataset,
            Each ensemble reported is a span-voting ensemble combining all models in Table \ref{table:final_ensem}.
            "keep" subscript refers to keeping uninformative answer spans as discussed in \ref{subsec:uninfor},
            on the other hand, "remove" subscript points to removing uninformative answer spans.
            }
            \label{table:test}
        \end{center}
    \end{table*}

    \subsection{Development Phase}
    In this phase,  we did not have access to the test dataset.
    We trained our models on the training set as in \ref{subsec:training} while only saving the best performing model on the validation split.
    We observed a large variation (around $\pm3\%$) in the pRR score reported for the same starting checkpoint with different seeds, we relate this to the small size of the validation split.
    To enable fair comparison, we average the reported scores for the same model trained  multiple times with different seeds as shown in Table \ref{table:eval}.
    For the ensemble method, we considered 15 checkpoints with different seeds for each of AraBERT-v02$_\text{Large}$, AraBERT-v02$_\text{Base}$ and ARBERT (marked with $\dagger$ in the table), adding up to 45 experts, labelled as \textbf{Ensemble$_\textbf{Vanilla}$} in Table \ref{table:eval}.
    After that, we performed the post-processing step on the ensemble outputs referred to as \textbf{Ensemble$_\textbf{POST}$} in Table \ref{table:eval}.
    Also from the table, QARiB$_\text{Base}$ and MARBERT are performing worse on average, because of being primarily targeted for dialectal Arabic and social media text.

    \subsection{Competition Final Testing Phase}

        During the competition's final testing phase, we had access to the test dataset without labels, and participants were required to submit at most 3 submissions produced by their proposed systems.
        We trained our models on both the training and development datasets as in \ref{subsec:training}. The trained models were combined in an ensemble as discussed in \ref{subsec:ensemble},
        then we performed the post-processing.
        Table \ref{table:final_ensem} shows the number of models used as experts for span-voting ensemble in the test phase.
        In Table \ref{table:test}, we outline the official results obtained for our submissions.
        All of them are ensemble-based due to the significant variations we observed in the development phase.
        Combining \textbf{all} of the models in Table \ref{table:final_ensem} followed by post-processing the output predictions with uninformative span \textbf{removal} performs best,
        this is marked in the table as \textbf{Ensemble$_\text{remove}$}.
         \fontdimen2\font=.77ex
        There is a significant gap between the results in the development and test phases as in tables \ref{table:eval} and \ref{table:test} respectively,
        we relate this to the small size of the validation split against the test split,
        another reason is excluding 36 samples from the test split since their questions were similar to the public splits as indicated by the organizers in the official test phase results.
        \fontdimen2\font=\origiwspc
        \begin{table}[ht]
            \begin{center}
                \begin{tabular}{|l|P{1.9cm}|P{1.8cm}|P{1.3cm}|}
                    \hline
                    \textbf{Models}           & \textbf{Num} \\
                    \hline
                    arabertv02$_\text{Large}$ & 16           \\
                    \hline
                    arabertv02$_\text{Base}$  & 18           \\
                    \hline
                    ARBERT                    & 17           \\
                    \hline
                \end{tabular}
                \caption{Number of models involved in the final ensemble of the test phase.}
                \label{table:final_ensem}
            \end{center}
        \end{table}

    \section{Conclusion and Future Work}
    \label{sec:conc}
    In this work, we leveraged the pre-trained Arabic language models to solve the \quran QA 2022 Shared Task.
    We fine-tuned a variety of BERT models optimized for the Arabic language.
    We proposed some post-processing operations to enhance the quality of answers aligning with the official measure.
    Ensemble-based approaches are effective to produce more robust predictions.
    In the future, we will further study how to incorporate a stacking ensemble approach with multiple stages to achieve better performance
    rather than a voting ensemble as used in this study.
    We will also investigate why we observed huge variations in the reported results
    by performing extensive cross-validation.


    \section{Acknowledgements}
    We appreciate the efforts and assistance of Dr Moustafa Alzantot regarding the paper-writing phase,
    and recommendations during the implementation.

    \section{Bibliographical References}\label{reference}
    \label{main:ref}

    \bibliographystyle{lrec2022-bib}
    \bibliography{main.bib}

\end{document}